\documentclass[runningheads, envcountsame, a4paper]{llncs}

\usepackage{cite}
\usepackage{amsmath,amssymb,amsfonts}
\usepackage{textcomp}
\usepackage{xcolor}
\usepackage{bm}
\usepackage{subfigure}
\usepackage{booktabs}
\usepackage{stackengine}
\usepackage{algorithm}
\usepackage{algorithmicx}
\usepackage{algpseudocode}
\usepackage[misc]{ifsym}
\usepackage{multirow}
\usepackage{diagbox}
\usepackage{color}
\usepackage[hyphens]{url}
\usepackage[pdftex]{graphicx}
\usepackage{epstopdf}
\usepackage{microtype}
\usepackage{pdfpages}
\begin{document}
\title{Topic-to-Essay Generation with \\ Comprehensive Knowledge Enhancement}
\titlerunning{Topic-to-Essay Generation with Knowledge Enhancement}
%
\author{Zhiyue Liu \and
Jiahai Wang(\Letter) \and
Zhenghong Li}
\authorrunning{Z. Liu, J. Wang, and Z. Li}

\toctitle{Topic-to-Essay Generation with Comprehensive Knowledge Enhancement}
\tocauthor{Zhiyue~Liu, Jiahai~Wang, and Zhenghong~Li}

%
\institute{
School of Computer Science and Engineering, Sun Yat-sen University\\
Guangzhou, China\\
\email{$\{$liuzhy93,lizhh98$\}$@mail2.sysu.edu.cn, wangjiah@mail.sysu.edu.cn}}
\maketitle              
\begin{abstract}
Generating high-quality and diverse essays with a set of topics is a challenging task in natural language generation.
Since several given topics only provide limited source information, utilizing various topic-related knowledge is essential for improving essay generation performance. However, previous works cannot sufficiently use that knowledge to facilitate the generation procedure.
This paper aims to improve essay generation by extracting information from both internal and external knowledge. Thus, a topic-to-essay generation model with comprehensive knowledge enhancement, named TEGKE, is proposed. For internal knowledge enhancement, both topics and related essays are fed to a teacher network as source information. Then, informative features would be obtained from the teacher network and transferred to a student network which only takes topics as input but provides comparable information compared with the teacher network.
For external knowledge enhancement, a topic knowledge graph encoder is proposed. Unlike the previous works only using the nearest neighbors of topics in the commonsense base, our topic knowledge graph encoder could exploit more structural and semantic information of the commonsense knowledge graph to facilitate essay generation.
Moreover, the adversarial training based on the Wasserstein distance is proposed to improve generation quality. Experimental results demonstrate that TEGKE could achieve state-of-the-art performance on both automatic and human evaluation.

\keywords{Topic-to-essay generation \and Knowledge transfer \and Graph neural network \and Adversarial training.}
\end{abstract}
\section{Introduction}
\begin{figure*}[t]
\centering
\includegraphics[width=0.70\textwidth]{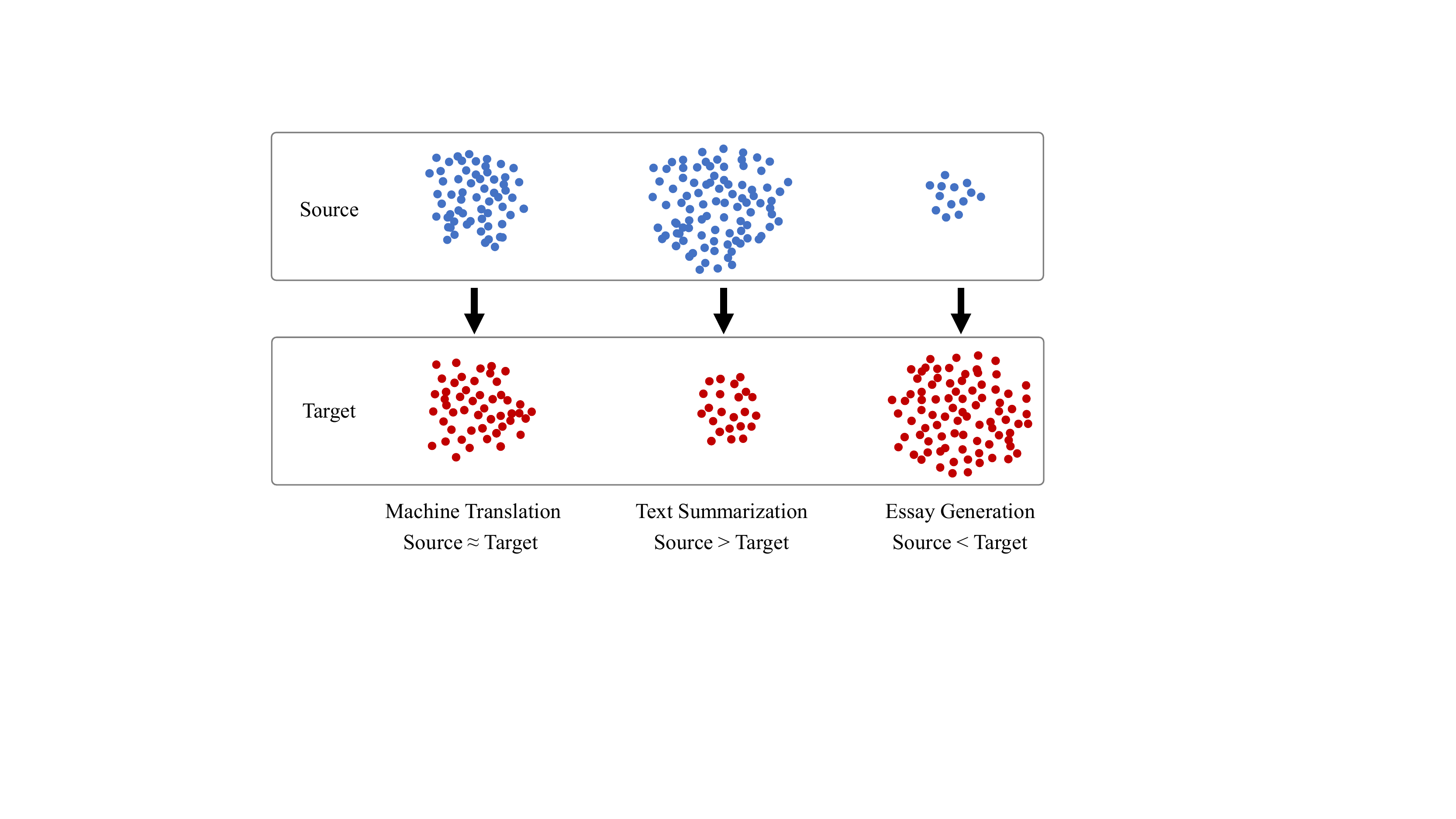}
\caption{Toy illustration of the information volume on different text generation tasks.} \label{fig0}
\end{figure*}

Topic-to-essay generation (TEG) is a challenging task in natural language generation, which aims at generating high-quality and diverse paragraph-level text under the theme of several given topics.
Automatic on-topic essay generation would bring benefits to many applications, such as news compilation~\cite{leppanen2017data}, story generation~\cite{fan2018hierarchical}, and intelligent education. Although some competitive results for TEG have been reported in the previous works using deep generative models~\cite{feng2018topic, yang2019enhancing, qiao2020sentiment}, the information gap between the source topic words and the targeted essay blocks their models from performing well. The comparison of information flow between TEG and other text generation tasks is illustrated in Fig.~\ref{fig0}~\cite{yang2019enhancing}. For machine translation and text summarization, the source provides enough information to generate the targeted text.
However, for the TEG task, the information provided by only the topic words is much less than that contained in the targeted text during generation, making the generated essays low-quality.

The proper utilization of various topic-related knowledge is essential to enrich the source information, which has not been sufficiently explored.
Incorporating the external knowledge from related common sense into the generation procedure is an efficient way to improve the TEG performance. However, in the commonsense knowledge graph, previous works~\cite{yang2019enhancing, qiao2020sentiment} only consider the nearest neighbor nodes of topic words, and neglect the multi-hop neighbors which would bring more structural and semantic information. Moreover, without considering external knowledge, their models cannot fully exploit the relation between topics and essays to assist the generation procedure.

This paper proposes a topic-to-essay generation model with comprehensive knowledge enhancement, named TEGKE. By extracting both internal and external knowledge, TEGKE greatly enriches the source information. Besides, the adversarial training based on the Wasserstein distance is proposed to further enhance our model. Thus, there are three key parts, including internal knowledge enhancement, external knowledge enhancement, and adversarial training.

For internal
knowledge enhancement, our model is based on the auto-encoder framework including a teacher network and a student network. Inspired by the conditional variational auto-encoder (CVAE) framework, the teacher network takes both topic words and related essays as source information to get informative latent features catching the high-level semantics of the relation between topics and essays. Then, a decoder could better reconstruct the targeted essay conditional on these features.
Since only topic words could be used as the input source during inference, the informative features (i.e., internal knowledge) from the teacher network would be transferred to the student network. Different from CVAE that trains the recognition network and the prior network to be close to each other in the latent space, the teacher network in TEGKE maintains an independent training procedure. Then, the student network is forced to be close to the teacher network. That is, the student could take only topics as input but output comparable informative latent features compared with the teacher.

For external knowledge enhancement, ConceptNet~\cite{speer2012representing} is employed as the commonsense knowledge base. Different from the previous works only using the nearest neighbors of topics, a topic-related knowledge graph is extracted from ConceptNet, which consists of multi-hop neighbors from the source topic words. Then, a topic knowledge graph encoder is proposed to perform on the multi-hop knowledge graph. It employs a compositional operation to obtain graph-aware node representations (i.e., external knowledge), which could conclude the structural information and the semantic information. The external knowledge is involved in the essay generation and helps select a proper decoding word.

Moreover, a discriminator is introduced for adversarial training. For alleviating the discrete output space problem of text, previous works adopt the adversarial training based on reinforcement learning (RL), which has the drawbacks of less-informative reward signals and high-variance gradients~\cite{Caccia2020Language}.
In contrast, this paper proposes to directly optimize the Wasserstein distance for the adversarial training, which avoids the problem of vanishing gradients and provides strong learning signals~\cite{gulrajani2017improved}. Based on the Wasserstein distance, the discriminator could operate on the continuous valued output instead of discrete text~\cite{subramanian2017adversarial}. For aligning essays with the related topics, topics are combined with generated essays as generated samples and combined with targeted essays as real samples. By the minimax game, the discriminator would provide an informative learning signal guiding our model to generate high-quality essays.

In summary, our contributions are as follows:
\begin{quote}
\begin{itemize}
\item A topic-to-essay generation model is proposed based on the knowledge transfer between a teacher network and a student network. The teacher network could obtain informative features for the student network to learn, making the student network provide abundant information with only topics as the input source.

\item A topic knowledge graph encoder is proposed to perform on the multi-hop knowledge graph extracted from the commonsense base. It helps our model exploit the structural and semantic information of the knowledge graph to facilitate essay generation. Moreover, a discriminator is introduced to improve generation quality by the adversarial training based on the Wasserstein distance.

\item Experimental results on both automatic evaluation and human evaluation demonstrate that TEGKE could achieve better performance than the state-of-the-art methods.
\end{itemize}
\end{quote}

\section{Related Work}
As a text generation task, TEG aims at generating high-quality and diverse paragraph-level text with given topics, which has drawn more attention.
This task is first proposed by Feng et al.~\cite{feng2018topic}, and they utilize the coverage vector to integrate topic information.
For enriching the input source information, external commonsense knowledge has been introduced for TEG~\cite{yang2019enhancing, qiao2020sentiment}. Besides, Qiao et al.~\cite{qiao2020sentiment} inject the sentiment labels into a generator for controlling the sentiment of a generated essay. However, during essay generation, previous works~\cite{yang2019enhancing, qiao2020sentiment} only consider the nearest neighbors of topic nodes in the commonsense knowledge graph. This limitation blocks their models from generating high-quality essays. For better essay generation, this paper makes the first attempt to utilize both structural and semantic information from the multi-hop knowledge graph.

Poetry generation is similar to TEG, which could be regarded as a generation task based on topics. A memory-augmented neural model is proposed to generate poetry by balancing the requirements of linguistic accordance and aesthetic innovation~\cite{zhang2017flexible}.
The CVAE framework is adopted with adversarial training to generate diverse poetry~\cite{li2018generating}. Yang et al.~\cite{yang2018generating} use hybrid decoders to generate Chinese poetry. RL algorithms are employed to improve the poetry diversity criteria~\cite{yi2018automatic} directly. Different from poetry generation showing obvious structured rules, the TEG task needs to generate a paragraph-level unstructured plain text, and such unstructured targeted output brings severe challenges for generation.

The RL-based adversarial training~\cite{yu2017seqgan, guo2018long} is used to improve essay quality in previous works~\cite{yang2019enhancing, qiao2020sentiment}. However, the noisy reward derived from the discriminator makes their models suffer from high-variance gradients. In contrast, our model directly optimizes the Wasserstein distance for the adversarial training without RL, achieving better generation performance.

\section{Methodology}
\subsection{Task Formulation}
Given a dataset including pairs of the topic words $\mathbf{x} = (x_1, ..., x_m)$ and the related essay $\mathbf{y} = (y_1, ...,y_n)$, for solving the TEG task, we want a $\theta$-parameterized model to learn each pair from the dataset and generate a coherent essay under given topic words, where the number of essay words $n$ is much larger than that of topic words $m$. Then, the task could be formulated as obtaining the optimal model with $\mathbf{\hat{\theta}}$ which maximizes the conditional probability as follows:
\begin{equation}
\mathbf{\hat{\theta}} = {\rm arg\ max}_{\mathbf{\theta}}P_{\theta}(\mathbf{y}|\mathbf{x}).
\end{equation}

\subsection{Model Description}
\begin{figure*}[t]
\centering
\includegraphics[width=0.965\textwidth]{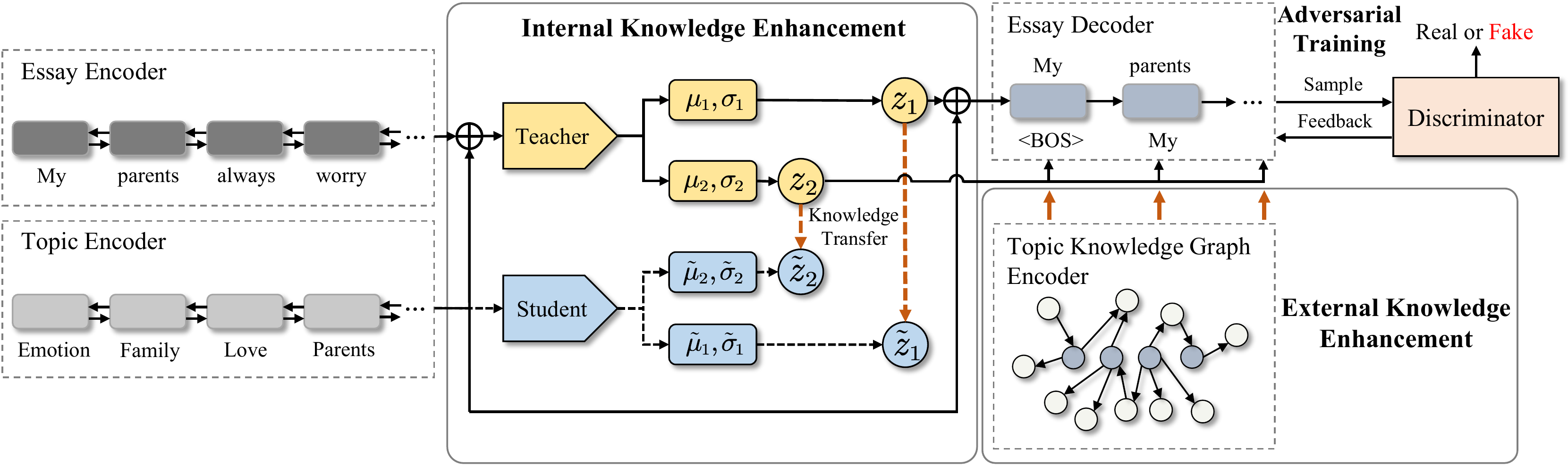}
\caption{Overview of the proposed model. Our model uses the teacher network for training (black solid arrows), and the student network for inference (black dotted arrows). The student network learns the latent features from the teacher network (red dotted arrows) for internal knowledge enhancement. The information from the topic knowledge graph encoder is integrated at each decoding step (red solid arrows) for external knowledge enhancement. During adversarial training, the generated essays are fed to the discriminator which provides learning signals as feedback.} \label{fig1}
\end{figure*}

Our TEGKE is based on the auto-encoder framework, utilizing both internal and external knowledge to enhance the generation performance. As shown in Fig.~\ref{fig1}, the model mainly contains three encoders (i.e., a topic encoder, an essay encoder, and a topic knowledge graph encoder) and an essay decoder. A discriminator is introduced at the end of the essay decoder for adversarial training.

For internal knowledge enhancement, the topic encoder and the essay encoder encode the topic words and the targeted essay sequence as $x_{\rm enc}$ and $y_{\rm enc}$, respectively. The latent features $z_1$ and $z_2$ are obtained from a teacher network  taking both $x_{\rm enc}$ and $y_{\rm enc}$ as input. Then, a student network, which takes $x_{\rm enc}$ solely as input, produces $\tilde{z}_1$ and $\tilde{z}_2$ to learn from $z_1$ and $z_2$ as internal knowledge, respectively. The essay decoder would generate a topic-related essay by receiving the latent features from the teacher network during training or those from the student network during inference.

For external knowledge enhancement, the multi-hop topic knowledge graph is constructed from the commonsense knowledge base, ConceptNet. Then, the topic knowledge graph encoder could represent the topic-related structural and semantic information as external knowledge to enrich the source information. The extracted external knowledge is attentively involved in each decoding step of the essay decoder to help select proper words and boost generation performance.

Through the adversarial training based on the Wasserstein distance, the discriminator could make the generated essay more similar to the targeted essay, which improves essay quality.

\subsubsection{Topic Encoder and Essay Encoder.}
The topic encoder employs a bidirectional gated recurrent unit (GRU)~\cite{cho2014learning}, which integrates the information of the topic sequence from both forward and backward directions. The topic encoder reads the embeddings of topic words $\mathbf{x}$ from both directions and obtains the hidden states for each topic word as follows:
\begin{equation}
\overrightarrow{h^x_{i}} = \overrightarrow{{\rm GRU}}(\overrightarrow{h^x_{i-1}}, e(x_i)),\  \overleftarrow{h^x_{i}} = \overleftarrow{{\rm GRU}}(\overleftarrow{h^x_{i+1}}, e(x_i)),
\end{equation}
where $e(x_i)$ is the embedding of $x_i$. The representation of the $i$-th topic is obtained as $h^x_i=[\overrightarrow{h^x_{i}}; \overleftarrow{h^x_{i}}]$, and ``$;$'' denotes the vector concatenation. The mean-pooling operation is conducted on the representations of all topics to represent $\mathbf{x}$ as $x_{\rm enc}={\rm mean}(h^x_1,...,h^x_m)$. Similarly, another bidirectional GRU is adopted as the essay encoder. The representation of the essay $\mathbf{y}$ could be obtained in the same way as the topic encoder does, which is denoted as $y_{\rm enc}$.
\paragraph{\underline{Internal Knowledge Enhancement}.}
Although the auto-encoder framework has shown competitive performance in many text generation tasks, the limited source information of the TEG task cannot provide sufficient information for the decoder to reconstruct the targeted output essay. This paper notices that informative latent features produced by the encoder are essential for a better decoding procedure. Inspired by the CVAE framework taking both the source and the target to train a recognition network, a teacher network is proposed by taking both the topics and essay as source information to get informative latent features for the essay decoder. Since only topics could be accessed during inference, a student network taking topic words solely as input is designed to learn from the teacher network's latent features as internal knowledge.
Different from CVAE that trains both the recognition network and the prior network to be close to each other in the latent space, the teacher network in our model maintains an independent training procedure following minimizing the reconstruction error. Because the teacher network is expected to provide strong supervision without being influenced by the student network.
The student network would generate latent features which learn the information from the teacher network's latent features through knowledge transfer. That is, the student network is pushed to be close to the teacher network in the latent space.

The teacher network consists of two feed-forward networks, and each network takes $x_{\rm enc}$ and $y_{\rm enc}$ as input to produce the mean and the diagonal covariance by two matrix multiplications, respectively. The latent features $z_1$ and $z_2$ are sampled from two Gaussian distributions defined by the above two feed-forward networks, respectively. During training, $z_1$ is used as a part of the essay decoder's initial hidden state, and $z_2$ is used as a part of the essay decoder's input at each step to provide more source information. The decoder receives $z_1$ and $z_2$ to optimize the training objective. Similarly, there are two feed-forward networks in the student network, where each network takes $x_{\rm enc}$ solely as input to sample a latent feature. Then, the student network's latent features $\tilde{z}_1$ and $\tilde{z}_2$ could be obtained. During inference, the decoder decodes $\tilde{z}_1$ and $\tilde{z}_2$ into a essay. Hence, above latent features could be obtained as follows:
\begin{align}
\begin{array}{c}
z_1 \sim \mathcal{N}(\mu_1,\sigma^2_1 \mathbf{I}) \\
z_2 \sim \mathcal{N}(\mu_2,\sigma^2_2 \mathbf{I})
\end{array},\  \left(\left[\begin{array}{c}
\mu_1 \\
\log \left(\sigma_1^{2}\right)
\end{array}\right],\left[\begin{array}{c}
\mu_2 \\
\log \left(\sigma_2^{2}\right)
\end{array}\right]\right)&={\rm Teacher}(x_{\rm enc},y_{\rm enc}), \\
\begin{array}{c}
\tilde{z}_1 \sim \mathcal{N}(\tilde{\mu}_1,\tilde{\sigma}^2_1 \mathbf{I}) \\
\tilde{z}_2 \sim \mathcal{N}(\tilde{\mu}_2,\tilde{\sigma}^2_2 \mathbf{I})
\end{array},\  \left(\left[\begin{array}{c}
\tilde{\mu}_1 \\
\log \left(\tilde{\sigma}^2_1\right)
\end{array}\right],\left[\begin{array}{c}
\tilde{\mu}_2 \\
\log \left(\tilde{\sigma}^2_2\right)
\end{array}\right]\right)&={\rm Student}(x_{\rm enc}),
\end{align}
where $\mathbf{I}$ is an identity matrix, and the reparametrization trick is used to sample the latent features. For enhancing the generation performance, the teacher network is trained to reconstruct the target, while the internal knowledge from the teacher network is transferred to the student network by minimizing the Kullback-Leibler (KL) divergence between the teacher's distributions and the student's distributions in the latent space as follows:
\begin{equation}
\mathcal{L}_{\rm trans} = {\rm KL}(\mathcal{N}(\tilde{\mu}_1, \tilde{\sigma}^2_1 \mathbf{I}) || \mathcal{N}(\mu_1, \sigma ^2_1 \mathbf{I})) + {\rm KL}(\mathcal{N}(\tilde{\mu}_2, \tilde{\sigma}^2_2 \mathbf{I}) || \mathcal{N}(\mu_2, \sigma ^2_2 \mathbf{I})).\label{Eq_KL}
\end{equation}

\subsubsection{Topic Knowledge Graph Encoder.}
Incorporating external commonsense knowledge is important to bridge the information gap between the source and the target.
Unlike previous works only considering the nearest neighbor nodes of topics, this paper constructs a topic knowledge graph queried by the topic words over a few hops from ConceptNet to assist the generation procedure. Then, a topic knowledge graph $\mathbf{G}=(\mathbf{V},\mathbf{R},\mathbf{E})$ could be obtained, where $\mathbf{V}$ denotes the set of vertices, $\mathbf{R}$ is the set of relations, and $\mathbf{E}$ represents the set of edges. The topic knowledge graph encoder is designed to integrate the topic-related information from $\mathbf{G}$.
By considering the topic knowledge graph, the objective of the TEG task could be modified as follows:
\begin{equation}
\mathbf{\hat{\theta}} = {\rm arg\ max}_{\mathbf{\theta}}P_{\theta}(\mathbf{y}|\mathbf{x, G}).
\end{equation}

\paragraph{\underline{External Knowledge Enhancement}.}
Appropriate usage of the structural and semantic information in the external knowledge graph plays a vital role in the TEG task.
Each edge $(u,r,v)$ in $\mathbf{G}$ means that the relation $r \in \mathbf{R}$ exists from a node $u$ to a node $v$. This paper extends $(u,r,v)$ with its reversed link $(v,r_{\rm rev},u)$ to allow the information in a directed edge to flow along both directions~\cite{marcheggiani2017encoding}, where $r_{\rm rev}$ denotes the reversed relation. For instance, given the edge $(worry,isa,emotion)$, the reversed edge $(emotion,isa{\_}r,worry)$ is added in $\mathbf{G}$. Our topic knowledge graph encoder is based on the graph neural network (GNN) framework, which could aggregate the graph-structured information of a node from its neighbors. Specifically, a graph convolution network (GCN) with $L$ layers is employed. For jointly embedding both nodes and relations in the topic knowledge graph, this paper follows Vashishth et al.~\cite{vashishth2019composition} to perform a non-parametric compositional operation $\phi$ for combining the neighbor node and the relation of a central node.
As shown in Fig.~\ref{fig2}, for a node $v \in \mathbf{V}$, its embedding would be updated at the $l$+1-th layer by aggregating information from its neighbors $N(v)$. The topic knowledge graph encoder treats incoming edges and outgoing edges differently to sufficiently encode structural information. Specifically, the related edges of the node $v$ could be divided into the set of incoming edges and that of outgoing edges, denoted as $\mathbf{E}_{{\rm in}(v)}$ and $\mathbf{E}_{{\rm out}(v)}$, respectively. Then, the node embedding of $v$ could be updated as follows:
\begin{figure}[t]
\centering
\includegraphics[width=\textwidth]{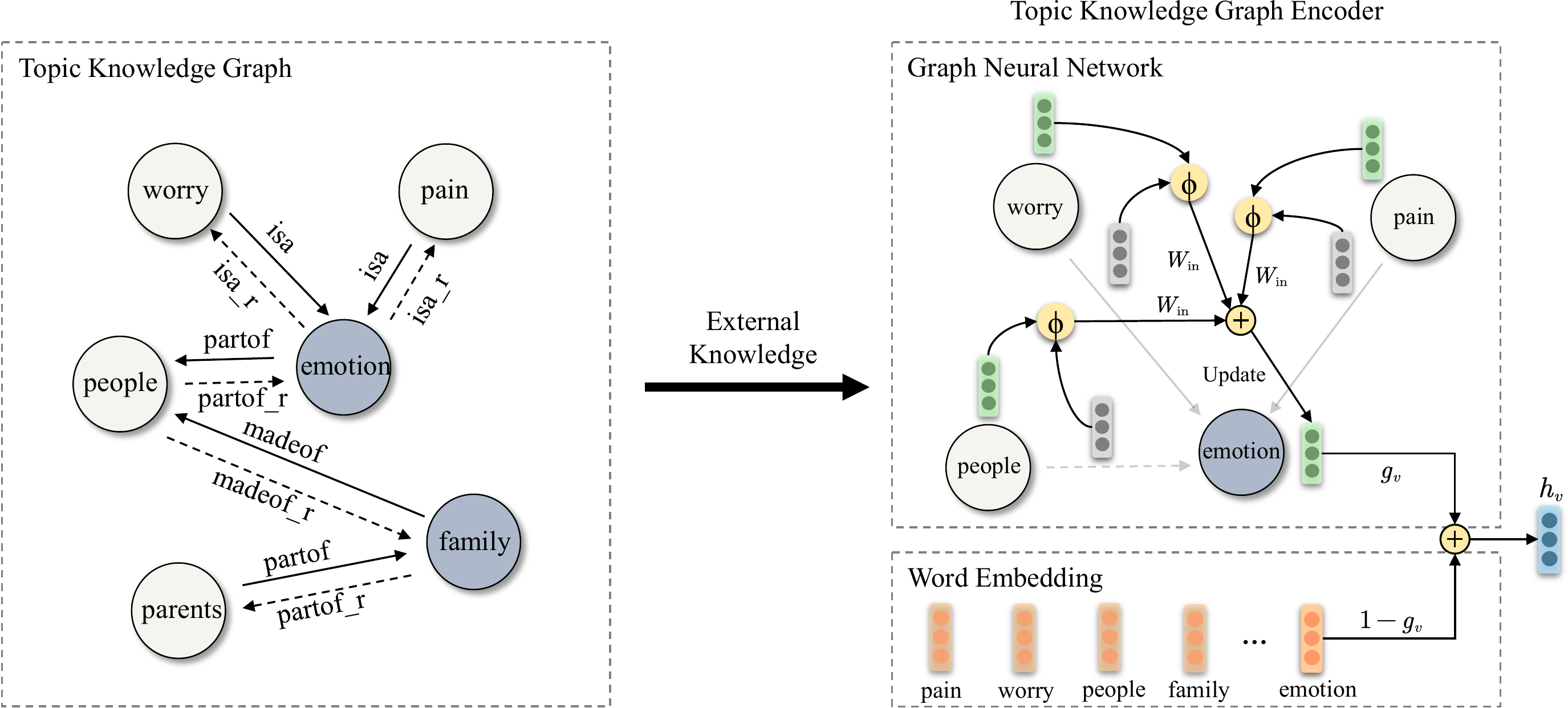}
\caption{Topic knowledge graph encoder. The graph neural network performs a compositional operation for a central node (e.g., emotion). Only incoming edges are considered in the diagram for clarity. The information from the topic knowledge graph is aggregated to update the embedding of the central node. Then, the final updated embedding is combined with the original word embedding to assist the essay decoding.} \label{fig2}
\end{figure}
\begin{align}
o^l_{v} &=\frac{1}{|N(v)|} \sum_{(u,r) \in N(v)}W^l_{{\rm dir}(r)} \phi(h^l_u, h^l_r), \\
h^{l+1}_v &={\rm ReLU}(o^l_v+W^l_{\rm loop}h^l_v),
\end{align}
where $h^0_{v}$ is initialized by the original word embedding, and $h^0_{r}$ is initialized by the relation embedding.
The weight matrix $W^l_{{\rm dir}(r)}$ is a relation-direction specific parameter at the $l$-th layer as follows:
\begin{equation}
W^l_{{\rm dir}(r)}=\left\{
\begin{aligned}
&W^l_{\rm in},  & (u,r,v) &\in \mathbf{E}_{{\rm in}(v)} \\
&W^l_{\rm out}, & (v,r,u) &\in \mathbf{E}_{{\rm out}(v)}
\end{aligned}
\right..
\end{equation}
The compositional operation employs $\phi(h^l_u, h^l_r)=h^l_u + h^l_r$ when incoming edges are considered, and $\phi(h^l_u, h^l_r)=h^l_u - h^l_r$ when outgoing edges are considered~\cite{vashishth2019composition}.
$o^l_v$ is the aggregated information from the $l$-th layer, and the weight matrix $W^l_{\rm loop}$ is used to transform $v$'s own information from the $l$-th layer. For the relation $r$, its embedding is updated as follows:
\begin{equation}
\begin{aligned}
h^{l+1}_r=W^l_{r}h^l_{r},
\end{aligned}
\end{equation}
where $W^l_r$ is a weight matrix. A gate mechanism is designed to combine $h^{L}_v$ containing graph knowledge and $h^0_v$ containing original semantic knowledge by:
\begin{align}
g_v &={\rm Sigmoid}(W_{\rm gate}[h^{L}_v;h^0_v]), \\
h_v &=g_v \odot h^{L}_v + (1-g_v) \odot h^0_v,
\end{align}
where $W_{\rm gate}$ is a weight matrix, and $\odot$ is the element-wise multiplication.

Finally, the node embedding $h_v$ is obtained to encode both structural and semantic information of the knowledge graph as external knowledge, involved in each decoding step for better essay generation.
\subsubsection{Essay Decoder.}
The essay decoder employs a single layer GRU. The initial hidden state $s_0$ is set with $s_0=[x_{\rm enc};z_1]$ containing the topics' representation and the latent feature. Both internal and external knowledge should be involved in each decoding step. Specifically, the hidden state $s_t$ of the decoder at time step $t$ is obtained as follows:
\begin{equation}
s_t = {\rm GRU}(s_{t-1}, [e(y_{t-1}); z_2; c^x_t; c^g_t]),
\end{equation}
where $e(y_{t-1})$ is the embedding of the essay word $y_{t-1}$ at the time step $t-1$,
$c^x_t$ is the topic context vector at the time step $t$, which integrates the output representations from the topic encoder by the attention mechanism as follows:
\begin{equation}
e^x_{t,i} =(\tanh(W_x s_{t-1} + b_x))^T h^x_i,\ \alpha ^x _{t,i} = \frac{\exp(e^x_{t,i})}{\sum ^m _{j=1} \exp(e^x_{t,j})},\ c^x_t = \sum ^m _{i=1} \alpha ^x _{t,i} h^x_i,
\end{equation}
and $c^g_t$ is the graph context vector, which integrates the representations of the graph nodes from the topic knowledge graph encoder as follows:
\begin{equation}
e^g_{t,v} =(\tanh(W_g s_{t-1} + b_g))^T h_v,\ \alpha ^g _{t,v} = \frac{\exp(e^g_{t,v})}{\sum _{u \in \mathbf{V}} \exp(e^g_{t,u})},\ c^g_t = \sum _{v \in \mathbf{V}} \alpha ^g _{t,v} h_v. \label{Eq_ATT}
\end{equation}

The internal knowledge from the latent feature $z_2$, and the external knowledge from the graph context vector $c^g_t$ would help the decoder select a proper word. Note that $z_1$ and $z_z$ would be replaced with $\tilde{z}_1$ and $\tilde{z}_2$ during inference.
Since our model takes both $\mathbf{x}$ and $\mathbf{y}$ as input when using the teacher network, the probability of obtaining an essay word for training is obtained by:
\begin{equation}
P_{\theta}(y_t|y_{<t},\mathbf{x}, \mathbf{y},\mathbf{G}) = {\rm Softmax}(W_o s_t + b_o).
\end{equation}
\subsubsection{Discriminator.}
A $\psi$-parameterized CNN-based discriminator~\cite{kim2014convolutional} $D_{\psi}$ is introduced in our model for adversarial training which would improve essay quality.
\paragraph{\underline{Adversarial Training}.} Due to the discrete output space problem of text generation, previous works heavily rely on the RL-based adversarial training which has less-informative reward signals and high-variance gradients.
In contrast, this paper proposes the adversarial training through the Wasserstein distance for TEG. Based on the Wasserstein distance, the discriminator could operate on continuous valued output and provide strong learning signals by distinguishing between a real text sequence of one-hot vectors and a generated text sequence of probabilities. Specifically, the hidden state $s_t$ of the essay decoder is employed to generate a probability output $y^{\theta}_{t}={\rm Softmax}(W_o s_t + b_o)$. Then, a sequence of outputs $\mathbf{y_{\theta}} = (y^{\theta}_1, ...,y^{\theta}_n)$ could be regarded as the generated essay for adversarial training. For aligning the generated essay with the related topics, the pair of the topics $\mathbf{x}$ and the ground truth essay $\mathbf{y}$ is fed to $D_{\psi}$ as the real sample, while the pair of $\mathbf{x}$ and $\mathbf{y_{\theta}}$ is treated as the generated sample. Then, the adversarial training objective based on the Wasserstein distance for $D_{\psi}$ is formulated by:
\begin{equation}
\mathcal{L}_{D_{\psi}} = D_{\psi}(\mathbf{x}, \mathbf{y_{\theta}}) - D_{\psi}(\mathbf{x},\mathbf{y}) + \lambda (||\nabla _{\hat{\mathbf{y}}}D_{\psi}(\mathbf{x}, \hat{\mathbf{y}})||_2 - 1)^2, \label{Eq_D}
\end{equation}
where the gradient penalty $(||\nabla _{\hat{\mathbf{y}}}D_{\psi}(\mathbf{x}, \hat{\mathbf{y}})||_2 - 1)^2$ weighted by $\lambda$ is imposed on the discriminator to enforce the Lipschitz constraint, and $\hat{\mathbf{y}}=\alpha \mathbf{y} + (1-\alpha)\mathbf{y_{\theta}}$ with $\alpha \sim {\rm Uniform}(0,1)$.
The auto-encoder framework in our model could act as a generator to minimize the following adversarial training objective as:
\begin{equation}
\mathcal{L}_{\rm adv} = - D_{\psi}(\mathbf{x}, \mathbf{y_{\theta}}) - \beta \log[P_{\theta}(\mathbf{y}|\mathbf{x, y, G})], \label{Eq_G}
\end{equation}
where the log-likelihood term $\log[P_{\theta}(\mathbf{y}|\mathbf{x, y, G})]$ weighted by $\beta$ would help align the generated essay with the topics further and keep generation diversity. The generator and the discriminator $D_{\psi}$ are alternately trained to play a minimax game, where $D_{\psi}$ assists the generator to obtain high-quality essays.
\subsection{Training and Inference}
For the training procedure, the latent features for decoding an essay are computed by the teacher network. Two training stages are employed in TEGKE. At the first training stage, the negative log-likelihood is minimized to reconstruct the ground truth essay $\mathbf{y} = (y_1, ...,y_n)$ as follows:
\begin{equation}
\mathcal{L}_{\rm rec} = \sum ^{n}_{t=1} -{\log}[P_{\theta}(y_t|y_{<t},\mathbf{x}, \mathbf{y},\mathbf{G})],
\end{equation}
where all parameters except the student network's parameters are optimized in an end-to-end manner. For transferring internal knowledge from the teacher network to the student network, the KL divergence between the student's distributions and the teacher's distributions is minimized by $\mathcal{L}_{\rm trans}$ of Eq.~(\ref{Eq_KL}) to optimize the student network's parameters.

At the second training stage, the auto-encoder framework in our model acts as a generator which is trained by $\mathcal{L}_{\rm adv}$ of Eq.~(\ref{Eq_G}). The discriminator is trained by $\mathcal{L}_{D_{\psi}}$ of Eq.~(\ref{Eq_D}) to provide a learning signal for the generator. Note that the student network is still optimized by $\mathcal{L}_{\rm trans}$ during the second stage.
For the inference procedure, the latent features for decoding are computed by the student network. The input to our model is the topics $\mathbf{x}$ and the topic knowledge graph $\mathbf{G}$, and then the decoder would generate a related essay. The pseudo code of TEGKE is shown in the supplementary material.

\section{Experiments}

\subsection{Datasets}
Experiments are conducted on the ZHIHU corpus~\cite{feng2018topic} consisting of real-world Chinese topic and essay pairs. The number of topic words is between 1 and 5. The length of an essay is between 50 and 100. For extracting external knowledge sufficiently, this paper constructs the topic knowledge graph from ConceptNet over 5 hops, and then 40 nodes are reserved per hop~\cite{ji2020language}. The topic knowledge graph is a subgraph of ConceptNet.
For the topic knowledge graph, the maximum number of nodes is 205, and the maximum number of edges is 912.
The training set and the test set contain 27,000 samples and 2,500 samples, respectively. We set $10\%$ of training samples as the validation set for hyper-parameters tuning. The experiments on the ESSAY corpus~\cite{feng2018topic} are shown in the supplementary material.

\subsection{Settings}
The essay decoder is a GRU with a hidden size of 1024. Both the topic encoder and the essay encoder are implemented as a bidirectional GRU with a hidden size of 512. The size of latent features is 512 in the teacher network and the student network. For the discriminator, the weight $\lambda$ of the gradient penalty is set to 10. The weight $\beta$ is set to 10. The vocabulary size is 50,000, and the batch size is set to 32. The 200-dim pretrained word embeddings~\cite{song2018directional} are shared by topics, essays, and initial graph nodes. The 200-dim randomly initialized vectors are used as initial graph relation embeddings. Adam optimizer~\cite{kingma2014adam} is used to train the model with the learning rate $10^{-3}$ for the first training stage, and the learning rate $10^{-4}$ for the second training stage.
\subsection{Baselines}
\textbf{TAV}~\cite{feng2018topic} encodes topic semantics as the average of the topic's embeddings and then uses an LSTM as a decoder to generate each word.\\
\textbf{TAT}~\cite{feng2018topic} enhances the decoder of TAV with the attention mechanism to select the relevant topics at each step.\\
\textbf{MTA}~\cite{feng2018topic} extends the attention mechanism of TAT with a topic coverage vector to guarantee that every single topic is expressed by the decoder.\\
\textbf{CTEG}~\cite{yang2019enhancing} introduces commonsense knowledge into the generation procedure and employs adversarial training to improve generation performance.\\
\textbf{SCTKG}~\cite{qiao2020sentiment} extends CTEG with the topic graph attention and injects the sentiment labels to control the sentiment of the generated essay. The SCTKG model without sentiment information is considered as a baseline, since the original TEG task does not take the sentiment of ground truth essays as input.
\subsection{Evaluation Metrics}
In this paper, both automatic evaluation and human evaluation are adopted to evaluate the generated essays.
\subsubsection{Automatic Evaluation.} Following previous works~\cite{feng2018topic,yang2019enhancing,qiao2020sentiment}, there are several automatic metrics considered to evaluate the model performance.\\
\textbf{BLEU}~\cite{papineni2002bleu}: The BLUE score is widely used in text generation tasks (e.g., dialogue generation and machine translation). It could measure the generated essays' quality by computing the overlapping rate between the generated essays and the ground truth essays.\\
\textbf{Dist-1, Dist-2}~\cite{li2016diversity}: The Dist-1 and Dist-2 scores are the proportion of distinct unigrams and bigrams in the generated essays, respectively, which measure the diversity of the generated essays.\\
\textbf{Novelty}~\cite{yang2019enhancing}: The novelty is calculated by the difference between the generated essay and the ground truth essays with similar topics in the training set. A higher score means more novel essays would be generated under similar topics.
\subsubsection{Human Evaluation.}
Following previous works~\cite{yang2019enhancing,qiao2020sentiment}, in order to evaluate the generated essays more comprehensively, 200 samples are collected from different models for human evaluation. Each sample contains the input topics and the generated essay. All 3 annotators are required to score the generated essays from 1 to 5 in terms of four criteria: \textbf{Topic-Consistency} (\textbf{T-Con.}), \textbf{Novelty} (\textbf{Nov.}), \textbf{Essay-Diversity} (\textbf{E-div.}), and \textbf{Fluency} (\textbf{Flu.}). For novelty, the TF-IDF features of topic words are used to retrieve the 10 most similar training samples to provide references for the annotators. Finally, each model's score on a criterion is calculated by averaging the scores of three annotators.
\begin{table*}[t]
\caption{Automatic and human evaluation results. $\uparrow$ means higher is better. $^*$ indicates statistically significant improvements ({\em p} $<0.001$) over the best baseline.}\label{tab1}
\centering
\resizebox{\columnwidth}{!}{
\begin{tabular}{l|c c cc|cccc}
\toprule
\multirow{2.5}{*}{Method} &\multicolumn{4}{c|}{Automatic Evaluation}&\multicolumn{4}{c}{Human Evaluation}\\
\cmidrule{2-9}
  & \textbf{BLEU($\uparrow$)~} & \textbf{Novelty($\uparrow$)~} & \textbf{Dist-1($\uparrow$)~} & \textbf{Dist-2($\uparrow$)~} &\textbf{T-Con.($\uparrow$)~}&\textbf{Nov.($\uparrow$)~}&\textbf{E-div.($\uparrow$)~}&\textbf{Flu.($\uparrow$)~} \\
\midrule
TAV    & 6.05   & 70.32& 2.69& 14.25 & 2.32&2.19&2.58&2.76\\
TAT    & 6.32   & 68.77& 2.25&  12.17 &1.76 &2.07 &2.32&2.93 \\
MTA    & 7.09   & 70.68& 2.24& 11.70  &3.14 & 2.87 & 2.17 &3.25\\
CTEG   & 9.72   & 75.71& 5.19& 20.49  &3.74  &3.34 &3.08&3.59\\
SCTKG  & 9.97 & 78.32& \textbf{5.73}& 23.16 & 3.89 &3.35 &3.90 &3.71\\
\midrule
TEGKE~~~ & \textbf{10.75$^*$} & \textbf{80.18$^*$}& 5.58& \textbf{28.11$^*$} & \textbf{4.12$^*$} &\textbf{3.57$^*$} &\textbf{4.08$^*$} &\textbf{3.82$^*$}\\
\bottomrule
\end{tabular}
}
\end{table*}
\subsection{Experimental Results}
\subsubsection{Automatic Evaluation Results.} The automatic evaluation results over generated essays are shown in the left block of Table~\ref{tab1}.
Compared with TAV, TAT, and MTA, TEGKE consistently achieves better results on all metrics. This illustrates that, without introducing sufficient knowledge, their models obtain unsatisfactory performance due to the limited source information. CTEG and SCTKG consider the nearest neighbor nodes of topics from ConceptNet as external information. In contrast, the multi-hop topic knowledge graph provides more structural and semantic information which is extracted by our topic knowledge graph encoder. Thus, our model outperforms the best baseline by 0.78 on the BLEU score, demonstrating that the potential of our model to generate high-quality essays.
Moreover, TEGKE could obtain competitive results on the Dist-1 scores, while greatly improving the Dist-2 and novelty scores by 4.95 and 1.86 over SCTKG, respectively. That is, the essays generated from our model would be more diverse and different from the essays in the training corpus.
In general, by integrating various internal and external knowledge into generation, TEGKE could achieve better quality and diversity simultaneously.

\subsubsection{Human Evaluation Results.} The human evaluation results are shown in the right block of Table~\ref{tab1}, and TEGKE could obtain the best performance on all metrics. The external knowledge incorporated by the topic knowledge graph encoder would help the decoder select topic-related words, and the adversarial training could further align generated essays with related topics.
Thus, our model outperforms the best baseline by 0.23 on the topic-consistency score, showing that the generated essays are more closely related to the given topics. The improvement over the novelty, essay-diversity, and fluency scores demonstrates that TEGKE could obtain better samples in terms of quality and diversity. This conclusion is similar to that drawn from the automatic evaluation.

\begin{table*}[t]
\caption{Ablation study results.}\label{tab2}
\centering
\resizebox{0.75\columnwidth}{!}{
\begin{tabular}{l|cccc}
\toprule
Method & \textbf{BLEU($\uparrow$)~} & \textbf{Novelty($\uparrow$)~} & \textbf{Dist-1($\uparrow$)~} & \textbf{Dist-2($\uparrow$)~} \\
\midrule
TEGKE & \textbf{10.75} & \textbf{80.18}& 5.58& 28.11 \\
TEGKE w/o EX &10.18 &78.67 &5.38&21.16\\
TEGKE w/o AD  & 10.63   & 80.09&\textbf{5.65}& \textbf{28.33}\\
TEGKE w/o EX \& AD~~~  & 9.78   & 79.42&5.46& 21.30 \\
\bottomrule
\end{tabular}
}
\end{table*}
\subsubsection{Ablation Study.}
To illustrate the effectiveness of our model's key parts, this paper performs an ablation study on three ablated variants: TEGKE without external knowledge enhancement (TEGKE w/o EX), TEGKE without adversarial training (TEGKE w/o AD), and TEGKE with only internal knowledge enhancement (TEGKE w/o EX \& AD). The results are shown in Table~\ref{tab2}.
\paragraph{\underline{Internal Knowledge Enhancement}.} Based on only the internal knowledge from the teacher network, TEGKE w/o EX \& AD achieves the worst results among variants. However, its performance is still comparable to CTEG adopting both adversarial training and commonsense knowledge, showing that the latent features produced by the TEGKE w/o EX \& AD's topic encoder benefit essay generation. Specially, TEGKE w/o EX \& AD increases Dist-1 by 0.27 and Dist-2 by 0.81 over CTEG.
This improvement comes from the teacher and student networks' various outputs, because our decoder generates essays depending on two latent features sampled from different Gaussian distributions.
The above results illustrate that utilizing a student to learn from a teacher makes our model learn the relation between topics and essays better, which enhances the model performance.
\paragraph{\underline{External Knowledge Enhancement}.} Compared with TEGKE, TEGKE w/o EX shows much inferior performance on all metrics. Specifically, TEGKE w/o EX drops 0.57 on the BLEU score, since the external knowledge would help the model select a topic-related word by exploring the topic words and their neighbors in the multi-hop topic knowledge graph. Besides, the diversity of generated essays from TEGKE w/o EX degrades, which is shown by the decline on the novelty, Dist-1, and Dist-2 scores. Specially, TEGKE w/o EX greatly drops 6.95 on Dist-2, due to lacking the commonsense knowledge to provide background information and enrich the input source. By utilizing external knowledge, TEGKE w/o AD still outperforms SCTKG on most metrics. That is, our graph encoder could extract more informative knowledge from the multi-hop knowledge graph.
\paragraph{\underline{Adversarial Training}.} Based on the adversarial training, TEGKE w/o EX boosts the BLEU score by 0.4 over TEGKE w/o EX \& AD, and only slightly sacrifices the novelty, Dist-1, and Dist-2 scores due to the inherent mode collapse problem in adversarial training. It demonstrates that the proposed adversarial training could effectively improve the essay quality. Compared with TEGKE w/o AD, TEGKE increases the BLEU score by 0.12, illustrating that our adversarial training could cooperate with the external knowledge enhancement. Since the external knowledge greatly enriches the source information and boosts the model performance, the improvement brought by the adversarial training is somewhat weakened when the topic knowledge graph is introduced.

\begin{figure*}[t]
\centering
\includegraphics[width=0.83\textwidth]{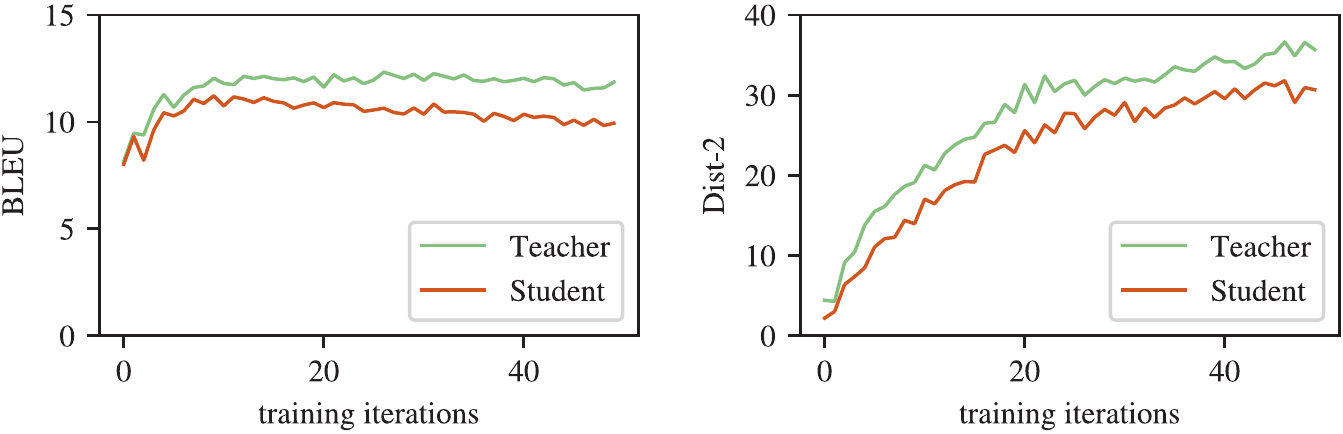}
\caption{Training curves. The BLEU score and the Dist-2 score are employed to measure quality and diversity, respectively. For both BLEU and Dist-2, the higher the better.} \label{fig3}
\end{figure*}
\subsection{Validity of Knowledge Transfer}
To illustrate the validity of transferring knowledge from the teacher network to the student network, the performance of our model using the teacher network and that using the student network is shown in Fig.~\ref{fig3}. The quality is measured by BLEU, and the diversity is measured by Dist-2. When our model uses the teacher network, the teacher network's latent features are fed to the decoder for generating essays. The model could maintain a stable training procedure and obtain excellent results since the ground truth essays are taken as input. The student network would learn from the teacher network's latent features. For the performance, the model using the student network closely follows that using the teacher network. Although the model using the student network performs slightly worse, the results on quality and diversity are still satisfactory.
\begin{figure*}[t]
\centering
\includegraphics[width=0.82\textwidth]{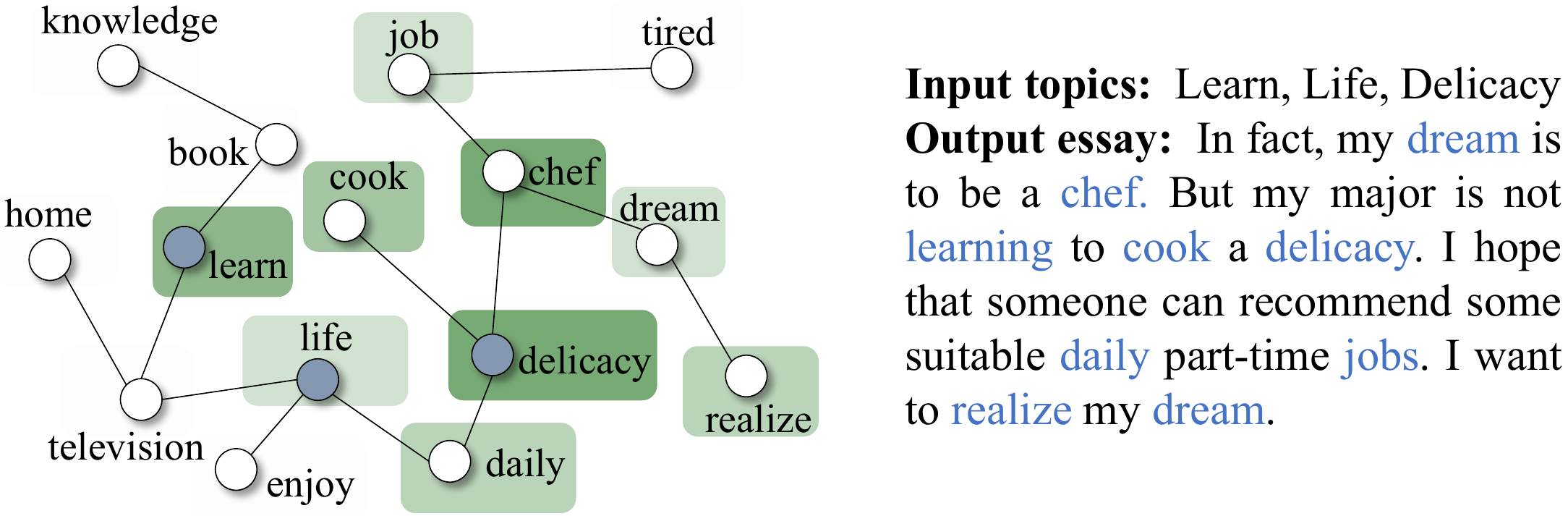}
\caption{Case study. The attention scores over the topic knowledge graph are shown on the left side. Deeper green indicates higher attention scores. The input topics and the generated essay are shown on the right side, where the selected words with higher attention scores are highlighted in blue. The original Chinese is translated into English.} \label{fig4}
\end{figure*}
\subsection{Case Study}
A case generated by our model is shown on the right side of Fig.~\ref{fig4}. Under the given topics ``learn'', ``life'', and ``delicacy'', TEGKE obtains a high-quality essay that mainly covers the semantics of input topics. The reason is that our model can integrate internal knowledge and abundant external knowledge into the generation procedure. By greatly enriching the source information, our model could generate novel and coherent essays.

To further illustrate the validity of our topic knowledge graph encoder, this paper visualizes the attention weights of Eq.~(\ref{Eq_ATT}) during the generation procedure on the left side of Fig.~\ref{fig4}. Compared with the previous works only considering the 1-hop neighbors of topics, our model could use the information from the multi-hop topic knowledge graph. For instance, in the generated essay, ``dream'' is a 2-hop neighbor of the topic ``delicacy'', and ``realize'' is a 3-hop neighbor of the topic ``delicacy''. It is observed that all nodes from the path (``delicacy'', ``chef'', ``dream'', and ``realize'') get higher attention scores during the generation procedure, indicating that the structural information of the graph is helpful. The generated essay is consistent with the topics' semantics since the topics ``learn'' and ``delicacy'' both obtain higher attention scores. Although the topic ``life'' does not appear in the generated essay, its 1-hop neighbor ``daily'' injects the corresponding semantic information about ``life'' into the generated essay.
\section{Conclusion}
This paper proposes a topic-to-essay generation model with comprehensive knowledge enhancement, named TEGKE. For internal knowledge enhancement, the teacher network is built by taking both topics and related essays as input to obtain informative features. The internal knowledge in these features is transferred to the student network for better essay generation. For external knowledge enhancement, the topic knowledge graph encoder is proposed to extract both the structural and semantic information from commonsense knowledge, which significantly enriches the source information.
Moreover, the adversarial training based on the Wasserstein distance is introduced to improve generation quality further.
Experimental results on real-world corpora demonstrate that TEGKE outperforms the state-of-the-art methods.
\section*{Acknowledgements}
This work is supported by the National Key R\&D Program of China (2018AAA0101203), and the National Natural Science Foundation of China (62072483).

%
%
%

\newpage


\section*{Supplementary material}

# Supplementary Material

## 1 Pseudo Code

TEGKE has two training stages: the first stage for minimizing the reconstruction error and the second stage for the adversarial training, which are shown in Algorithm 1 and Algorithm 2, respectively. The experiment code would be made publicly available for further research.

---
**Algorithm 1** First training stage of TEGKE.
---
**Require:** Encoders (including topic encoder, essay encoder, topic knowledge graph encoder, teacher network, and student network); essay decoder; training corpus $\{(\mathbf{x}, \mathbf{y})\}$; number of training iterations $N$; number of batches $B$.
**Ensure:** Encoders; essay decoder
1: Initialize encoders and essay decoder with random parameters $\theta$
2: **for** $n = 1, \cdots, N$ **do**
3:     **for** $b = 1, \cdots, B$ **do**
4:         Encode $\mathbf{x}$ and $\mathbf{y}$ into $x_{\text{enc}}$ and $y_{\text{enc}}$, respectively
5:         Sample latent features $z_1$ and $z_2$ from teacher network via Eq. (3)
6:         Compute node embeddings $h_v$ via Eqs. (7)-(12)
7:         Decode probabilities of generated essays via Eqs. (13)-(16)
8:         Update parameters $\theta$ expect student network's parameters via Eq. (19)
9:         Update student network's parameters via Eq. (5)
10:     **end for**
11: **end for**
12: **return** Encoders; essay decoder
---

---

**Algorithm 2** Second training stage of TEGKE.

---
**Require:** Generator (including topic encoder, essay encoder, topic knowledge graph encoder, teacher network, student network, and essay decoder); discriminator $D_\psi$; training corpus $\{(\mathbf{x}, \mathbf{y})\}$; number of training iterations $N_{\text{adv}}$; generator's updating steps $N_G$; discriminator's updating steps $N_D$.
**Ensure:** Generator
 1: Initialize $D_\psi$ with random parameters $\psi$
 2: Load pre-trained generator's parameters $\theta$ from first training stage
 3: **for** $n_{\text{adv}} = 1, \cdots, N_{\text{adv}}$ **do**
 4:     **for** $n_G = 1, \cdots, N_G$ **do**           ▷ Generator Training
 5:         Sample a batch of $(\mathbf{x}, \mathbf{y})$
 6:         Generate essay sequences of probabilities $\mathbf{y}^\theta$
 7:         Combine $\mathbf{x}$ and $\mathbf{y}^\theta$ as generated samples
 8:         Update parameters $\theta$ expect student network's parameters via Eq. (18)
 9:         Update student network's parameters via Eq. (5)
10:     **end for**
11:     **for** $n_D = 1, \cdots, N_D$ **do**           ▷ Discriminator Training
12:         Sample a batch of $(\mathbf{x}, \mathbf{y})$
13:         Generate essay sequences of probabilities $\mathbf{y}^\theta$
14:         Combine $\mathbf{x}$ and $\mathbf{y}$ as real samples
15:         Combine $\mathbf{x}$ and $\mathbf{y}^\theta$ as generated samples
16:         Update discriminator's parameters $\psi$ via Eq. (17)
17:     **end for**
18: **end for**
19: **return** Generator

## 2  Experimental Results on the ESSAY Corpus

Additional experiments are conducted on the ESSAY corpus. The number of topic words is 5. The length of an essay is between 50 and 120. The topic knowledge graph is constructed from ConceptNet over 5 hops, and 40 nodes are reserved per hop. The maximum number of nodes is 205, and the maximum number of edges is 954. After our preprocessing, the training set and the test set contain 300,000 samples and 5,000 samples, respectively. We set 10% of training samples as the validation set for hyper-parameters tuning.

The automatic evaluation results are shown in Table A1. Note that CTEG and SCTKG do not report the results on the ESSAY corpus, and the RL-based adversarial training would make their training time extremely long on ESSAY which has more topics and training samples than the ZHIHU corpus. In contrast, through the adversarial training based on the Wasserstein distance, our model gets rid of the complicated learning strategy and has reasonable training time.

Compared with TAV, TAT, and MTA, TEGKE boosts all metrics. It demonstrates that, without introducing sufficient knowledge, baselines obtain unsatisfactory performance due to the limited source information. By integrating various internal and external knowledge, TEGKE generates essays with better quality and diversity simultaneously. This conclusion is similar to that drawn from the automatic evaluation on the ZHIHU corpus.

**Table A1.** Automatic evaluation results. ↑ means higher is better. * indicates statistically significant improvements ($p < 0.001$) over the best baseline.

| Method | BLEU(↑) | Novelty(↑) | Dist-1(↑) | Dist-2(↑) |
|---|---|---|---|---|
| TAV | 3.47 | 80.38 | 1.46 | 7.35 |
| TAT | 4.32 | 78.80 | 2.14 | 11.21 |
| MTA | 4.78 | 78.53 | 1.54 | 9.33 |
| TEGKE | **7.46*** | **81.12*** | **4.54*** | **34.86*** |

## 3    Effectiveness of the Multi-Hop Topic Knowledge Graph

Different numbers of hops constructing the topic knowledge graph would influence the performance. For the external knowledge enhancement of TEGKE, the results in Fig. A1 show the influence of different numbers of hops on the ZHIHU corpus. The increase of the hop number triggers the increase of the topic knowledge graph's size and the improvement of the model performance (i.e., both BLEU and Dist-2 increase). However, a bigger knowledge graph would consume more graphics card memory. Compared with the topic knowledge graph based on 1-hop neighbors, the topic knowledge graph based on 5-hop neighbors significantly increases the BLEU and Dist-2 scores. According to the memory capacity of our graphics card, we utilize 5-hop neighbors to construct the topic knowledge graph.

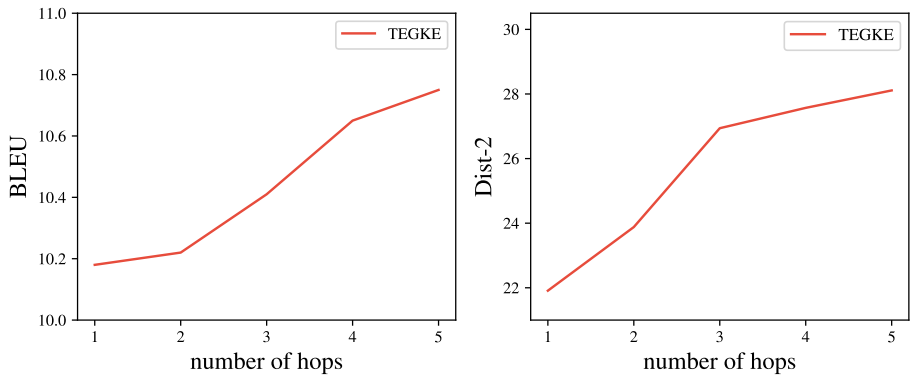

**Fig. A1.** Influence of numbers of hops constructing the topic knowledge graph. The BLEU score and the Dist-2 score are employed to measure quality and diversity, respectively. For both BLEU and Dist-2, the higher the better.


\end{document}